\documentclass{article}
\usepackage[utf8]{inputenc}
\usepackage{graphicx}
\usepackage[english]{babel}
\usepackage{tikz}
\usepackage{color}
\usepackage{amsthm}
\usepackage{bm}
\usepackage{amsmath}
\usepackage{amsmath,amssymb}
\usepackage{graphicx}
\usepackage[colorinlistoftodos]{todonotes}
\usepackage[colorlinks=true, allcolors=blue]{hyperref}
\usepackage[T1]{fontenc}
\usepackage{tikz}

\usepackage[a4paper,
            bindingoffset=0.2in,
            left=1in,
            right=1in,
            top=1in,
            bottom=1in,
            footskip=.25in]{geometry}

\usepackage{epsfig}
\usepackage{amsmath}
\usepackage{amssymb}
\usepackage{booktabs}
\usepackage{xcolor,colortbl}
\usepackage{multirow}
\usepackage{subcaption}

\usepackage[capitalize]{cleveref}
\crefname{section}{Sec.}{Secs.}
\Crefname{section}{Section}{Sections}
\Crefname{table}{Table}{Tables}
\crefname{table}{Tab.}{Tabs.}

\def\attn{CSA}
\def\meth{MS\attn}

\def\short{MS\attn}

\def\X{\textbf{X}}
\def\Q{\textbf{Q}}
\def\K{\textbf{K}}
\def\V{\textbf{V}}
\def\W{\textbf{W}}
\def\R{\mathbb{R}}

\def\eg{\textit{e.g.}}
\def\ie{\textit{i.e.}}

\newcommand\red[1]{\textcolor{black}{#1}}

\definecolor{mygray}{gray}{0.9}
\newcommand\grayl[1]{\multicolumn{1}{>{\columncolor{mygray}}l}{#1}}
\newcommand\grayc[1]{\multicolumn{1}{>{\columncolor{mygray}}c}{#1}}

\usepackage{algorithm}
\usepackage{algorithmic}

\title{Vision Backbone Enhancement via Multi-Stage Cross-Scale Attention}
\author{Liang Shang\textsuperscript{\rm 1}, Yanli Liu\textsuperscript{\rm 2}, Zhengyang Lou\textsuperscript{\rm 1}, Shuxue Quan\textsuperscript{\rm 2}, Nagesh Adluru\textsuperscript{\rm 1}, \\Bochen Guan\textsuperscript{\rm 2}, and William A. Sethares\textsuperscript{\rm 1}
}

\date{\today}

\begin{document}

\maketitle

\footnotetext[1]{University of Wisconsin - Madison, Madison, WI, US. lshang6@wisc.edu, zlou4@wisc.edu, adluru@wisc.edu, sethares@wisc.edu}
\footnotetext[2]{OPPO US Research Center, Palo Alto, CA, US. yanli.liu@oppo.com, quanshuxue@gmail.com,\\ bochen.guan@oppo.com}

\begin{abstract}
     Convolutional neural networks (CNNs) and vision transformers (ViTs) have achieved remarkable success in various vision tasks. However, many architectures do not consider interactions between feature maps from different stages and scales, which may limit their performance. In this work, we propose a simple add-on attention module to overcome these limitations via multi-stage and cross-scale interactions. Specifically, the proposed Multi-Stage Cross-Scale Attention (MSCSA) module takes feature maps from different stages to enable multi-stage interactions and achieves cross-scale interactions by computing self-attention at different scales based on the multi-stage feature maps. Our experiments on several downstream tasks show that MSCSA provides a significant performance boost with modest additional FLOPs and runtime.
     
\end{abstract}

\section{Introduction}

\begin{figure*}
    \centering
    \includegraphics[scale = 0.55]{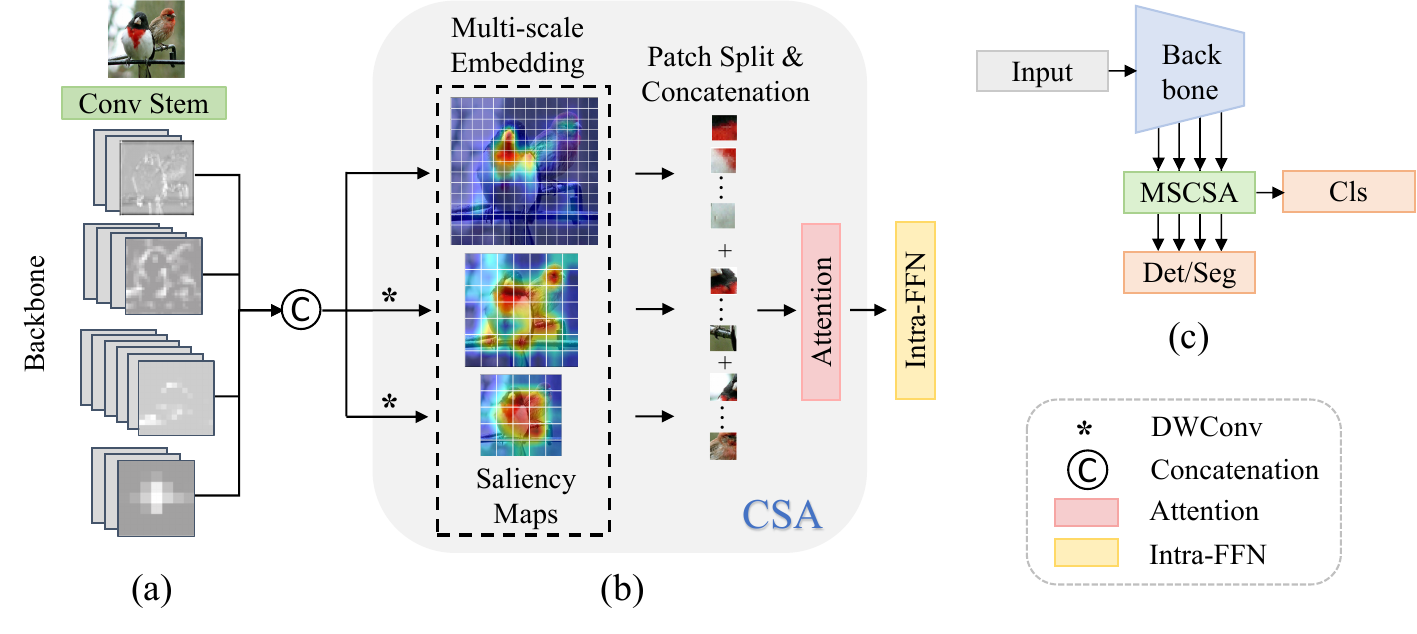}
    \caption{(a) \textbf{Feature aggregation for multi-stage interaction.} In \meth, the output feature maps from different stages of the backbone are downsampled and concatenated along the channel dimension to enable multi-stage interaction.
    (b) \textbf{Cross-scale attention (\attn) and Intra-FFN.} 
    \attn~first applies depthwise convolutions to generate key and value tensors at different scales, then concatenates them to compute an attention map across scales. Intra-FFN only carries out FFN computation within each stage to save FLOPs.
    (c) \textbf{\meth~as an add-on module for vision backbones.} \meth~is designed as an add-on module that takes feature maps from different stages of the vision backbone as input. The output of \meth~can be used for various downstream tasks such as classification, detection, and segmentation.} 
    \label{fig:overview}
\end{figure*}

Convolutional neural networks (CNNs) and vision transformers (ViTs) have been widely used as backbones in various vision tasks, such as image classification~\cite{cnn2017,vit2020}, object detection~\cite{maskrcnn2017,resnet2016,swin2021}, semantic segmentation~\cite{deeplab2017,vgg2014,segformer2021}, video recognition~\cite{stconv2014,mvit2021}, and medical imaging~\cite{liu2019fully,vitmedical2022}. However, many existing architectures do not fully exploit interactions between feature maps from different stages and different scales, which may restrict their performance on downstream tasks.

For example, in object detection, the connections between \red{smaller and larger parts of an object} and objects of different sizes are essential for accurate localization. Similarly, in semantic segmentation, the correlations between representative and abstract feature maps at different stages of a network can be critical for accurate predictions. However, these interactions may be missed in traditional single-path top-to-bottom architectures.

The first type of connection can be incorporated by allowing interactions between image patches of different scales. We call this {\em cross-scale interaction}. For CNNs, shortcut connections~\cite{resnet2016,densenet2017} and other architectural topologies~\cite{googlenet2015,inceptionv32016,inceptionv42017,res2net2019,hrnet2020}, namely different ways of connecting feature aggregation modules, can be used to explore the interaction between feature maps in different scales. Similarly, ViTs also try to address this issue in various ways. For instance,   \cite{coat2021,hrformer2021,hrvit2022,crossvit2021,mpvit2022,lgformer2021} extend the idea of cross-scale interactions to several network topologies for fusing feature maps. Others~\cite{patchformer2022,crossformer2021,dpt2021,p2t2022,focal2021,lvt2022} propose modifying existing modules to introduce cross-scale interactions.

The correlations between representative and abstract feature maps can be modeled by allowing the feature maps to interact between different stages of the backbone. We call this {\em multi-stage interaction}.
To incorporate multi-stage interaction, CoaT~\cite{coat2021} and TopFormer~\cite{topformer2022} propose parallel branches to connect feature maps at different stages. However, these approaches have limitations. CoaT's Co-Scale mechanism is computationally complex and is limited to a specific backbone structure, while TopFormer only downsamples feature maps in the parallel branch, which limits interactions to multiple stages but not across scales.

In this work, we propose a simple add-on attention module, {\em Multi-Stage Cross-Scale Attention (\meth)}, to enhance both multi-stage and cross-scale interactions in CNN and ViT backbones, as illustrated in~\cref{fig:overview}. Our approach combines Cross-Scale Attention (\attn) and Intra-Feed-Forward Networks (Intra-FFN) to achieve these interactions with moderate additional computational cost. 

Specifically, to enable multi-stage interactions, we collect feature maps from different stages of the backbone and concatenate them as a multi-stage feature map, as shown in~\cref{fig:overview}a. To achieve cross-scale interactions, we first apply depthwise convolutions in \attn~to obtain embeddings at different scales, where an example of the saliency maps of these multi-scale feature embeddings are depicted in~\cref{fig:overview}b. These saliency maps illustrate that the features in coarser scales focus more on the overall objects of interest, while the features in finer scales focus on different parts of the birds, \eg, the beak in the top scale and the feet in the middle scale. These multi-scale embeddings are then concatenated and used to compute a cross-scale attention map to exploit information across multiple scales. Furthermore, we replace half of the usual feed-forward networks (FFNs) with the Intra-FFN, which only carries out intra-stage FFN computation and saves FLOPs in compensation for the additional computation in CSA. 

We conduct experiments on several CNN and ViT-based vision backbones, including TopFormer~\cite{topformer2022}, CoaT~\cite{coat2021}, ResNet~\cite{resnet2016}, PVTv2~\cite{pvtv22022}, Swin~\cite{swin2021}, and the state-of-the-art models P2T~\cite{p2t2022} and ConvNeXt~\cite{liu2022convnet} on common benchmark tasks to validate the effectiveness of \meth. Our results show that \meth~provides a significant performance boost on these backbones, especially on downstream tasks. Furthermore, this improvement only comes with modest increases in FLOPs and runtime.

Our contributions can be summarized as follows:
\begin{enumerate}
\itemsep = -0.1cm
    \item We propose Cross-Scale Attention (\attn) and Intra-Feed-Forward Networks (Intra-FFN) to enable multi-stage and cross-scale feature interactions in an efficient manner.
    \item We design Multi-Stage Cross-Scale Attention (\meth) as an add-on module that applies to a variety of vision backbones and leads to a clear performance boost.
    \item We demonstrate that \meth~brings a clear performance boost on various vision backbones. In most cases, this comes with less than $10\%$ additional FLOPs and runtime.
    
\end{enumerate}

\section{Related Works}

\noindent
\textbf{Vision Transformers (ViTs).} The introduction of vision transformers (ViTs) in~\cite{vit2020} generated significant interest, and standard training strategies are provided by DeiT~\cite{deit2021}. However, the quadratic complexity of the self-attention mechanism~\cite{attention2017} leads to a heavy computational burden for ViTs. To overcome this, Swin Transformer~\cite{swin2021} proposes computing the self-attention within local windows, a strategy that has been widely applied and developed in subsequent works~\cite{cswin2022,dw-vit2022,shuffletransformer2021,mixformer2022,hrvit2022,nat2022,dat2022,scalabelvit2022}. Other recent papers~\cite{glance2021,cat2021,crossformer2021,maxvit2022} aim to improve the performance further by managing the global interactions between windows. Additionally, some works~\cite{pvt2021,pvtv22022,mvit2021,mvitv22022,longformer2021,scalabelvit2022,p2t2022,twins2021,edgevit2022,focal2021} use feature down-sampling within the attention blocks to reduce the computational cost.

\noindent
\textbf{Multi-stage and cross-scale interaction in ViTs.} As the ViT research community grows, researchers have started exploring new ViT topology designs. CoaT~\cite{coat2021} and TopFormer~\cite{topformer2022} build parallel paths to add connections between feature maps in different stages, while CoaT further employs cross-scale interactions by keeping feature maps in different resolutions in the parallel path. CrossViT~\cite{crossvit2021} proposes a dual-path architecture of different patch sizes, where the cross-scale interactions occur through the classification (CLS) tokens~\cite{bert2018,vit2020,deit2021} when exchanging between the paths. Inspired by HRNet~\cite{hrnet2020}, HRFormer~\cite{hrformer2021} and HRViT~\cite{hrvit2022} maintain several feature paths of different resolutions to apply cross-scale interactions. LG-Transformer~\cite{lgformer2021} and MPViT~\cite{mpvit2022} generate multiple paths of varying resolutions in the single-stage block structure and add cross-scale interactions within the stage. 

In contrast, many other works focus on modifying specific modules within the ViT architecture to achieve cross-scale interactions. CrossFormer~\cite{crossformer2021}, DPT~\cite{dpt2021}, and PatchFormer~\cite{patchformer2022} generate multi-scale feature maps in patch embedding layers, while Focal~\cite{focal2021}, LVT~\cite{lvt2022}, Shunted~\cite{shunted2022}, and P2T~\cite{p2t2022} suggest using multi-scale feature embeddings for key and value tensors or query tensors in the attention blocks.

\noindent
\textbf{Comparison to previous works.} \meth~applies a tailored attention mechanism in a parallel branch to achieve effective multi-stage and cross-scale interactions. In contrast, CoaT~\cite{coat2021} only uses simple operations such as addition and pooling/upsampling to fuse feature maps from different scales, while other models mainly focus on one of the multi-stage and cross-scale interactions. 

Moreover, \meth~applies to popular CNN and ViT backbones. In comparison, many existing cross-scale and multi-stage interactions are designed for certain backbones. For instance, HRViT~\cite{hrvit2022} and HRFormer~\cite{hrformer2021} inherit the HRNet~\cite{hrnet2020} structure, which makes them not directly applicable to other backbones. The Pooling-based MHSA in P2T~\cite{p2t2022} and the Shunted Self-Attention~\cite{shunted2022} maintains multi-scale embedding in their attention block, which makes them not directly applicable to window-based attentions with global interactions~\cite{glance2021,cat2021,crossformer2021,maxvit2022}. To the best of our knowledge, \meth~is the first add-on module that boosts the performance of vision backbones with efficient multi-stage cross-scale interactions.

\section{Multi-Stage Cross-Scale Attention Design}


\subsection{Overview}

Multi-Stage Cross-Scale Attention (\cref{fig:mscsa_1}) is a lightweight add-on module capable of incorporating multi-stage and cross-scale interactions. It is designed to boost the performance of popular vision backbones with only a modest increase in computation (approximately $10\%$ more FLOPs). To achieve efficient multi-stage interactions, \meth~first collects feature maps from different stages of a vision backbone, downsamples, and combines them along the channel dimension to obtain a multi-stage feature map. Then, cross-scale interactions are performed by Cross-Scale Attention (\attn), as in \cref{fig:attn}, where the key and value tensors contain information from feature maps of different scales. To allow for the additional computational cost in \meth, we replace half of the Feed-Forward Networks (FFNs) after the \attn~with an Intra-Feed-Forward Network (Intra-FFN), where the comparison between FFN and Intra-FFN is illustrated in~\cref{fig:ffn}.

This section introduces the \attn~mechanism and Intra-FFN in~\cref{sec:csa,sec:intra-ffn}, respectively. Several architectural variants of \meth~designed for different tasks are discussed in~\cref{sec:variants}. 
The examples of applying \meth~to various vision backbones can be found in the Appendix.

\subsection{Cross-Scale Attention (\attn)}
\label{sec:csa}

\begin{figure}[t]
    \centering \includegraphics[width=0.65\linewidth]{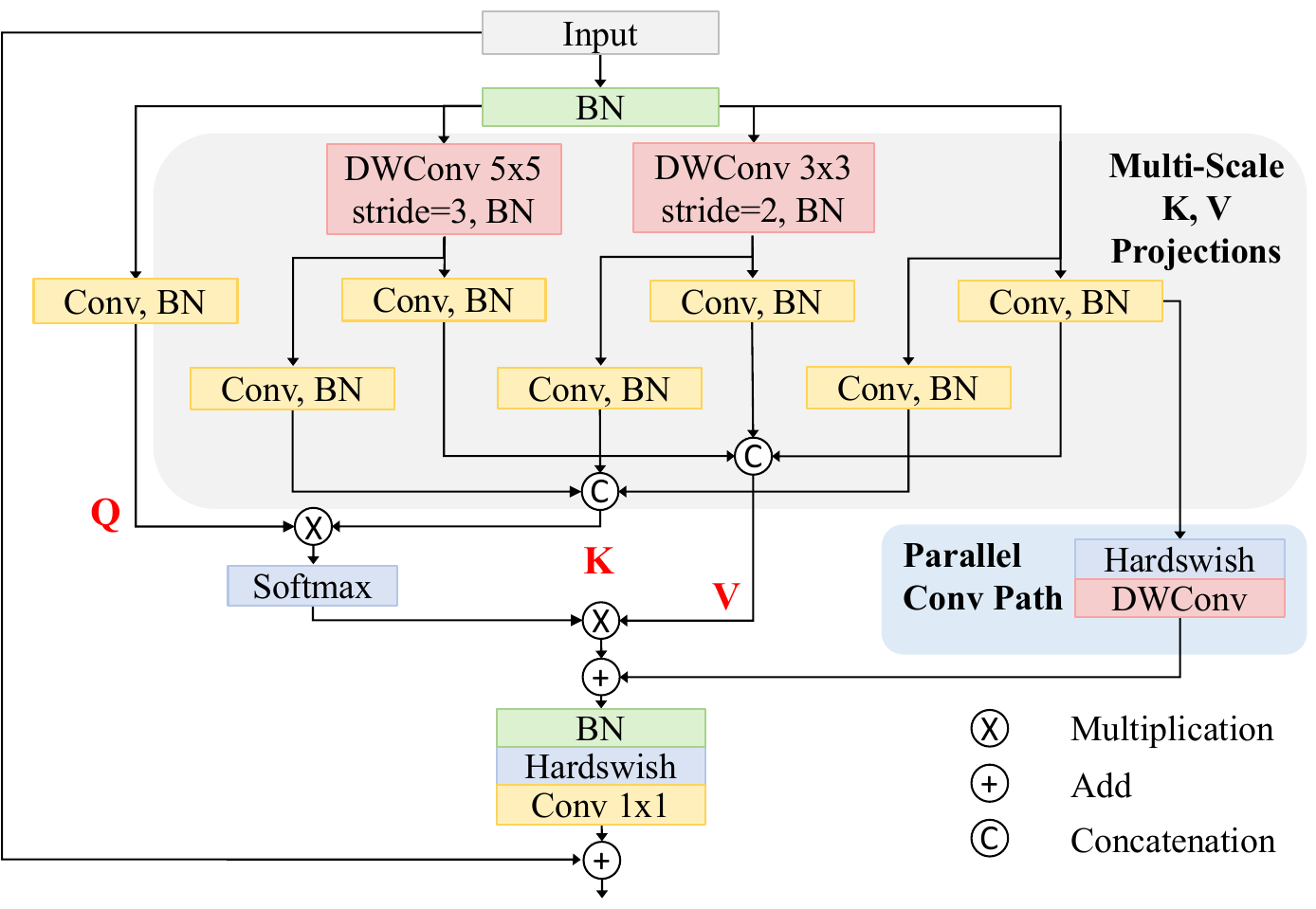}
    \caption{\textbf{Cross-Scale Attention.} The \attn~consists of Q projection, Multi-Scale K, V Projections (MSP), and a Parallel Convolution Path (PCP)~\cite{hrvit2022}. The multi-scale K, V projections generate the key, value tensors in three different scales, while the parallel convolution path acts as the relative position encoding.}
    \label{fig:attn}
\end{figure}

CSA aims to perform interactions across coarse and fine scales, \red{it introduces interactions between feature maps of different scales, which is desirable for downstream tasks}. In contrast to commonly used multi-head self-attention~\cite{attention2017,vit2020,deit2021} mechanisms, our \attn~uses Multi-Scale key, value tensor Projections (MSP), and a Parallel Convolution Path (PCP)~\cite{hrvit2022}, as illustrated in~\cref{fig:attn}. 

\noindent
\textbf{Multi-Scale key, value tensor Projections.} A natural way to incorporate cross-scale interactions in self-attention is to use multi-scale feature maps in the attention mechanism. 
Here, we introduce the multi-scale projections for key and value tensors, where the combined key and value tensors consist of feature maps from the original scale and two downsampled scales. When computing the attention map, every token, \ie, image patch, in the query tensor can query information in the key tensor from three different scales. The output of the self-attention is then a weighted sum of features from different scales. 

Specifically, from the input feature map $\X \in \R^{hw \times c}$ where $h, w, c$ are the height, width, and channel dimension, respectively, the query tensor $\Q$ can be computed from a linear projection
\begin{equation}
    \Q = \X\W^{q} \in \R^{hw \times d}
    \label{eq:q_proj}
\end{equation}
following the original attention mechanism~\cite{attention2017} with $d$ being the head dimension. On the other hand, for the key tensor $\K$ and value tensor $\V$, the input $\X$ is first downsampled using two depthwise convolutional operators:
\begin{gather}
    \begin{aligned}
        \X_{0} &= \X 
        ,\\
        \X_{1} &= \text{DWConv}_{1}(\X) 
        ,\\
        \X_{2} &= \text{DWConv}_{2}(\X) 
        ,
    \end{aligned}
    \label{eq:downsample}
\end{gather}
where $\X_{i}$ has a size of $h_{i}w_{i} \times c$ for $i=0,1,2$ as
\begin{gather}
    \begin{aligned}
        h_{0} &= h, &\ &w_{0} = w, \\
        h_{1} &= \left\lfloor \frac{h - 1}{2} + 1 \right\rfloor, &\ &w_{1} = \left\lfloor \frac{w - 1}{2} + 1 \right\rfloor, \\
        h_{2} &= \left\lfloor \frac{h - 1}{3} + 1 \right\rfloor, &\ &w_{2} = \left\lfloor \frac{w - 1}{3} + 1 \right\rfloor,
    \end{aligned}
\end{gather}
and $\text{DWConv}_{1}$ is a depthwise convolution for $2\times$ downsampling, while $\text{DWConv}_{2}$ achieves $3\times$ downsampling. Afterwards, the feature maps in different scales $\X_0, \X_1, \X_2$ generate their key and value tensors via linear projections separately:
\begin{gather}
    \begin{aligned}
        \K_{0} &= \X_{0}\W^{k}_{0} 
        , &&\V_{0} = \X_{0}\W^{v}_{0} 
        ,\\
        \K_{1} &= \X_{1}\W^{k}_{1} 
        , &&\V_{1} = \X_{1}\W^{v}_{1} 
        ,\\
        \K_{2} &= \X_{2}\W^{k}_{2} 
        , &&\V_{2} = \X_{2}\W^{v}_{2} 
        ,
        \label{eq:kv_proj}
    \end{aligned}
\end{gather}
where $\K_{i}$ has a size of $h_{i}w_{i} \times d$ and $\V_{i}$ has a size of $h_{i}w_{i} \times 2d$ for $i=0,1,2$. Following the dimension settings of LeViT~\cite{levit2021}, the value tensors have twice the head dimension as the query and key tensors have. As $2d$ could be the commonly used head dimension in self-attention in other works, the computational cost of our attention is reduced. These key and value tensors representing different scales are then concatenated together to form the multi-scale key and value tensors
\begin{equation}
    \begin{aligned}
        \K &= \text{Concat}(\K_{0},\K_{1},\K_{2}) 
        ,\\
        \V &= \text{Concat}(\V_{0},\V_{1},\V_{2}) 
        ,
    \end{aligned}
    \label{eq:kv_concat}
\end{equation}
where the concatenated $\K$ and $\V$ are of $(hw + h_{1}w_{1} + h_{2}w_{2}) \times d$ and $(hw + h_{1}w_{1} + h_{2}w_{2}) \times 2d$, respectively.
Following the original attention mechanism, the query tensor $\Q$, multi-scale key and values tensors $\K, \V$ are sent to compute the output of self-attention
\begin{equation}
    \text{Attn}(\Q, \K, \V) = \text{Softmax}\left(\frac{\Q\K^{T}}{\sqrt{d}}\right)\V \in \R^{hw \times 2d}.
    \label{eq:attn}
\end{equation}
As the multi-scale $\K, \V$ are approximate of sizes $(hw + \frac{1}{4}hw + \frac{1}{9}hw) \times d$ and $(hw + \frac{1}{4}hw + \frac{1}{9}hw) \times 2d$, \attn~only adds marginal computation cost compared to its single-scale counterpart, whose $\K, \V$ have the sizes of $hw \times d$ and $hw \times 2d$, respectively. The saliency maps in~\cref{fig:overview}b help demonstrate the benefit of maintaining cross-scale interactions in CSA.

\noindent
\textbf{Parallel Convolution Path.} The self-attention mechanism in~\cref{eq:attn} is often considered poor a distinguishing between long-range and short-range relationships, where structural information~\cite{structinfo2020} and local relationships~\cite{localrelation1999} play a significant role in vision tasks. Accordingly, we incorporate the Parallel Convolution Path (PCP), a technique drawn from HRViT~\cite{hrvit2022}, which acts as a relative positional encoding to preserve local information in self-attention. The modified self-attention with PCP now becomes
\begin{gather}
    \begin{aligned}
        \text{ModifiedAttn}(\Q, \K, \V) = \text{Attn}(\Q, \K, \V) + \text{ConvPath}(\V_{0}) \in \R^{hw \times 2d}.
    \end{aligned}
    \label{eq:modified_attn}
\end{gather}
A detailed description of the PCP is in the Appendix.

\subsection{Intra-Feed-Forward Network (Intra-FFN)}
\label{sec:intra-ffn}

\begin{figure}[t]
    \centering
    \includegraphics[width=0.55\linewidth]{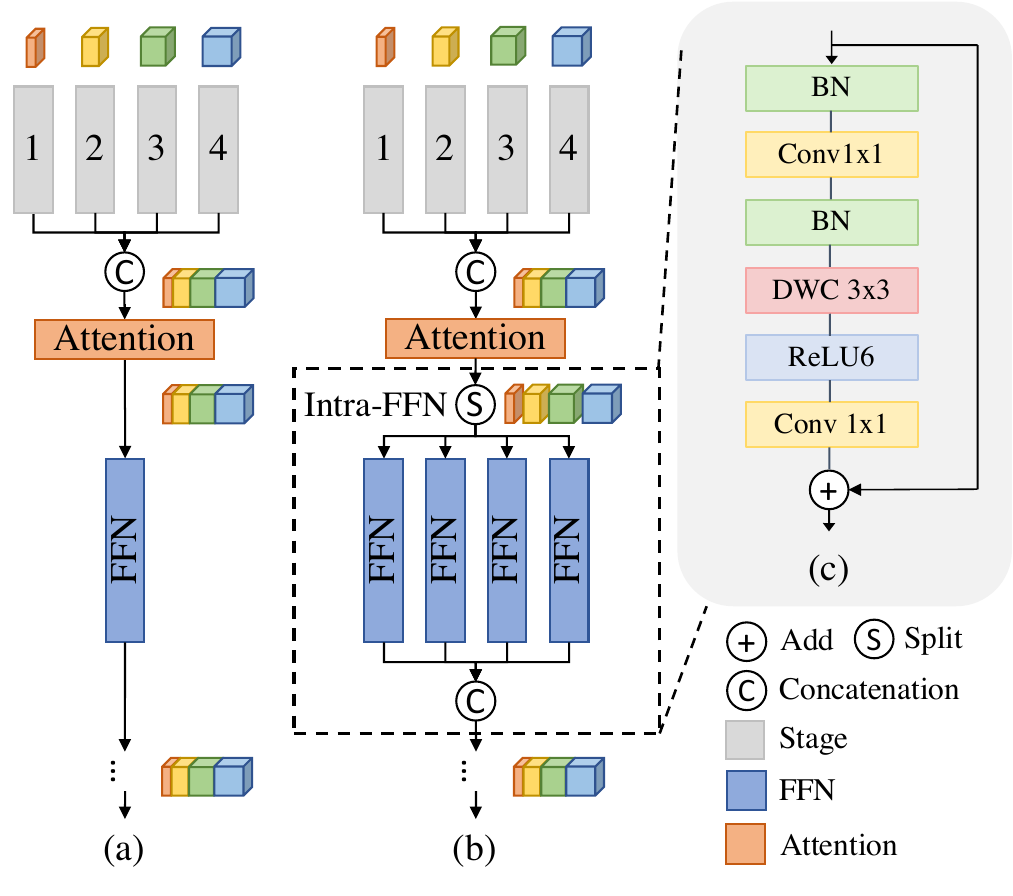}
    \caption{\textbf{FFN and Intra-FFN.} (a) The commonly used self-attention and FFN. (b) The self-attention followed by our proposed Intra-FFN. (c) The detailed architecture of FFN. In Intra-FFN, the feature map is first split into four parts corresponding to the channel dimension in each stage and then fed into four FFNs separately.}
    \label{fig:ffn}
\end{figure}

The Feed-Forward Network (FFN) is also an essential part of the design of attention blocks. For the FFN, we follow the design in~\cite{ceit2021,localvit2021,topformer2022,pvtv22022} which is a 3$\times$3 depthwise convolution inserted between two linear projections, as shown in~\cref{fig:ffn}c. The two linear projections, \ie, 1$\times$1 convolutions, enable feature interaction between stages, while the depthwise convolution acts as another positional encoding layer to supplement the PCP~\cite{hrvit2022} in \attn~and to encourage local feature aggregation. 

Since the input to the FFN is the multi-stage feature map which usually has a large channel dimension, the two linear projections in FFN carry a massive amount of FLOPs. As shown in~\cref{tab:summary_pvtv2_b1_mscsa}, in one \meth~block, the FLOPs of a single FFN layer is close to two \attn~layers. To overcome this issue, we propose our Intra-Feed-Forward Network (Intra-FFN). In the Intra-FFN, the multi-stage feature map is first split into several parts, where the number of parts matches the number of stages in the backbone, and each part has a channel dimension corresponding to its original stage. Then, the feature maps from each of the parts are fed into several FFNs separately, and their outputs are concatenated back into a multi-stage feature map. In the \meth~block, half of the FFN layers are replaced by Intra-FFN layers, which allows savings in computation. The saved computation by Intra-FFN can be spent on enlarging the channel dimension of the multi-stage feature map and increasing the expansion ratio of linear projections in FFN, which enriches multi-stage interactions. The ablation study in~\cref{sec:major_ablation} confirms that Intra-FFN in \meth~leads to a performance gain with a similar computational cost compared to FFN.

\begin{table}[]
    \centering
    \begin{tabular}{l|cc}
        \toprule
        Component 
        & FLOPs (G)
        & Percentage
        \\
        \midrule
        CSA
        & 0.049
        & 2.1\%
        \\
        FFN
        & 0.081
        & 3.5\%
        \\
        Intra-FFN
        & 0.030
        & 1.3\%
        \\
        \bottomrule
    \end{tabular}
    \caption{\textbf{FLOPs summary of \meth~components in PVTv2-B1+\meth.} 
    Assuming the input image has a size of 224$\times$224.}
    \label{tab:summary_pvtv2_b1_mscsa}
\end{table}

\subsection{\meth~and Architectural Variants}
\label{sec:variants}

\begin{figure}[h]
    \centering\includegraphics[width=0.55\linewidth]{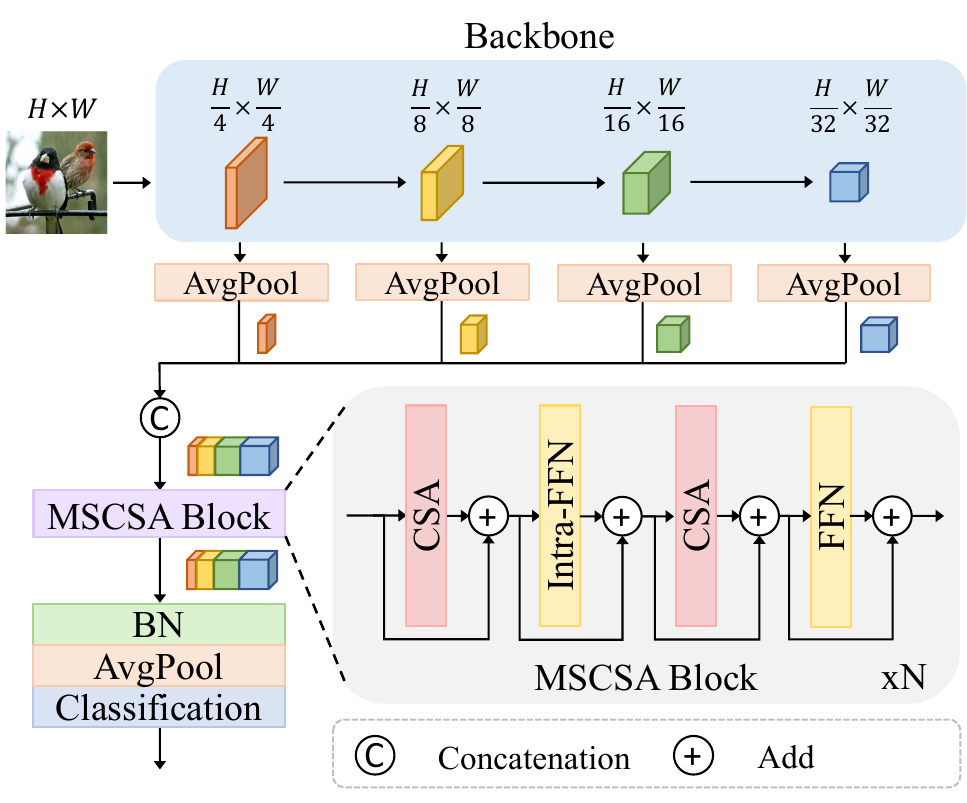}
    \caption{\textbf{Multi-Stage Cross-Scale Attention (\meth) for classification.} Feature maps from different scales are pooled (with averaging) to the same size. An additional 1$\times$1 convolution layer can be optionally applied to further decrease the number of channels and reduce FLOPs. Afterward, feature maps from all four stages are concatenated along the channel dimension and fed into several \meth~blocks.}
    \label{fig:mscsa_1}
\end{figure}

\begin{figure}[h]
    \centering
    \includegraphics[scale = 0.5]{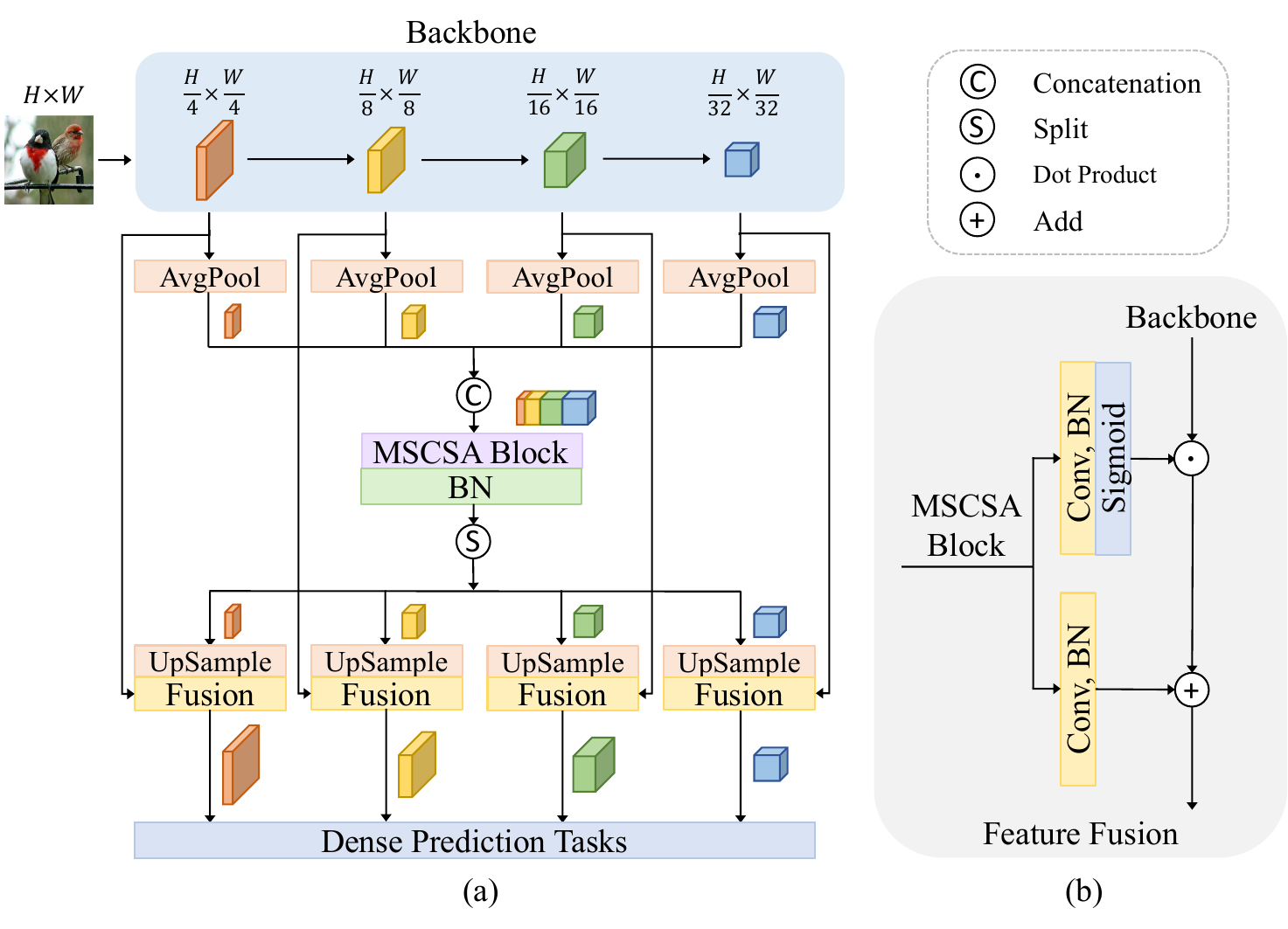}
    \hspace{-5mm}
    \caption{(a) \textbf{Multi-Stage Cross-Scale Attention (\meth) for dense prediction tasks.} For dense prediction tasks, the multi-stage feature map from \meth~blocks (same as in Fig. \ref{fig:mscsa_1}) is split into several parts corresponding to the number of channels in each stage and upsampled to their original resolutions. Together with the feature maps from the original backbone, they are fused through feature fusion layers. Finally, the fused feature maps in all scales are fed into downstream networks for dense prediction tasks. (b) \textbf{Detailed architecture of feature fusion layer.} The upsampled feature map from \meth~blocks generate injection weights and biases. Then, the weights and biases are fused with the feature map from the backbone via element-wise multiplication and addition, respectively.}
    \label{fig:mscsa_2}
\end{figure}

The construction of \meth~begins by taking inputs from feature maps in different stages of a vision backbone, where the backbone usually has four stages with resolutions from $H/4 \times W/4$ to $H/32 \times W/32$\footnote{Extending \meth~to vision backbones with different numbers of stages is straightforward.}. 
Feature maps from different scales are average-pooled to a target size, \eg, $H/32 \times W/32$, to enable efficient feature interactions. 
Moreover, for backbones with a large number of channels, an additional 1$\times$1 convolution layer will be applied to decrease the channel dimension and further reduce FLOPs. Afterward, feature maps from different stages are concatenated along the channel dimension to obtain a multi-stage feature map. This feature map is then fed into a stack of several \meth~blocks to apply feature interactions between different stages and scales, where the \meth~block is a stack of a \attn~layer followed by an Intra-FFN layer and another \attn~layer followed by an FFN layer. For classification and dense prediction tasks, we provide two architectural variants in Figs. \ref{fig:mscsa_1} and \ref{fig:mscsa_2}.

\section{Experiments}

To validate that \meth~can widely enhance the performance of vision backbones via multi-stage and cross-scale feature map interactions, we implement and evaluate \meth~on a variety of convolution and attention-based vision backbones. Our baselines include TopFormer~\cite{topformer2022} and CoaT~\cite{coat2021}. These two models have similar multi-stage feature interactions, where TopFormer has a convolution-based backbone, and CoaT has a ViT-based backbone. We substitute the attention modules and the parallel blocks in TopFormer and CoaT with \meth, respectively, and match the FLOPs of the corresponding baseline models. Moreover, we test our \meth~on CNN backbones ResNet~\cite{resnet2016} and ConvNeXt~\cite{liu2022convnet}, ViT backbones PVTv2~\cite{pvtv22022}, P2T~\cite{p2t2022}, and Swin~\cite{swin2021}.


The Cross-Scale Attention (CSA) in Sec. \ref{sec:csa} is designed to facilitate downstream tasks by introducing interactions between feature maps of different scales. Therefore, we mainly evaluate the performance of our \meth~integrated with these vision backbones on downstream tasks, including object detection and instance segmentation on the COCO dataset~\cite{coco2014} (\cref{sec:det}), and semantic segmentation on the ADE20K dataset~\cite{ade20k2017} (\cref{sec:seg}). For completeness, we also present the performance on image classification on the ImageNet-1k dataset~\cite{imnet2009} in the Appendix. Finally, we conduct a series of ablation studies in~\cref{sec:ablation} to test the effectiveness of each component of \meth.

\subsection{Object Detection and Instance Segmentation}
\label{sec:det}



\begin{table*}[t]
    \centering
    \resizebox{\textwidth}{!}{
    \begin{tabular}{c|l|cc|ccc|ccc}
        \toprule
        Method
        & Backbone
        & FLOPs (G)
        & FPS
        & $\text{AP}^{\text{b}}$
        & $\text{AP}^{\text{b}}_{50}$
        & $\text{AP}^{\text{b}}_{75}$
        & $\text{AP}^{\text{m}}$
        & $\text{AP}^{\text{m}}_{50}$
        & $\text{AP}^{\text{m}}_{75}$
        \\
        \midrule
        & CoaT Tiny$^{*}$
        & 260
        & 30
        & 42.7
        & 64.7
        & 46.5
        & 38.7
        & 61.7
        & 41.4
        \\
        & \grayl{\textbf{CoaT Tiny+\short}} \vline
        & \grayc{\textbf{237(-8.8\%)}} 
        & \grayc{\textbf{41(+36.7\%})} \vline
        & \grayc{\textbf{43.9}(+1.2)}
        & \grayc{66.0}
        & \grayc{48.1} \vline
        & \grayc{\textbf{39.8}(+1.1)}
        & \grayc{62.7}
        & \grayc{42.2}
        \\
        & CoaT Small
        & 428
        & 17
        & 46.5
        & 68.5
        & 51.1
        & 41.8
        & 65.5
        & 45.0
        \\
        & \grayl{\textbf{CoaT Small+\short}} \vline
        & \grayc{\textbf{318(-25.7\%)}} 
        & \grayc{\textbf{31(+82.4\%})} \vline
        & \grayc{46.4(-0.1)}
        & \grayc{69.2}
        & \grayc{50.7} \vline
        & \grayc{42.1(+0.3)}
        & \grayc{65.7}
        & \grayc{45.1}
        \\
        & ResNet-50$^{\dagger}$
        & 260
        & 103
        & 38.2
        & 59.0
        & 41.5
        & 35.4
        & 56.3
        & 37.9
        \\
        & \grayl{\textbf{ResNet-50+\short}} \vline
        & \grayc{271(+4.2\%)} 
        & \grayc{86(-18.4\%)} \vline
        & \grayc{\textbf{42.2}(+4.0)}
        & \grayc{63.8}
        & \grayc{46.1} \vline
        & \grayc{\textbf{38.7}(+3.3)}
        & \grayc{60.8}
        & \grayc{41.4}
        \\
        & PVTv2-B0
        & 196
        & 62
        & 38.2
        & 60.5
        & 40.7
        & 36.2
        & 57.8
        & 38.6
        \\
        \multirow{4}{*}{Mask R-CNN}
        & \grayl{\textbf{PVTv2-B0+\short}} \vline
        & \grayc{199(+1.5\%)} 
        & \grayc{59(-4.8\%)} \vline
        & \grayc{\textbf{41.0}(+2.8)}
        & \grayc{62.8}
        & \grayc{44.4} \vline
        & \grayc{\textbf{37.8}(+1.6)}
        & \grayc{59.9}
        & \grayc{40.7}
        \\
        & PVTv2-B1
        & 244
        & 47
        & 41.8
        & 64.3
        & 45.9
        & 38.8
        & 61.2
        & 41.6
        \\
        & \grayl{\textbf{PVTv2-B1+\short}} \vline
        & \grayc{250(+2.5\%)}  
        & \grayc{44(-6.4\%)} \vline
        & \grayc{\textbf{43.8}(+2.0)}
        & \grayc{65.6}
        & \grayc{47.9} \vline
        & \grayc{\textbf{40.0}(+1.2)}
        & \grayc{62.6}
        & \grayc{43.0}
        \\
        & PVTv2-B2
        & 309
        & 28
        & 45.3
        & 67.1
        & 49.6
        & 41.2
        & 64.2
        & 44.4
        \\
        & \grayl{\textbf{PVTv2-B2+\short}} \vline
        & \grayc{322(+4.2\%)} 
        & \grayc{27(-3.6\%)} \vline
        & \grayc{\textbf{46.0}(+0.7)}
        & \grayc{67.7}
        & \grayc{50.6} \vline
        & \grayc{\textbf{41.5}(+0.3)}
        & \grayc{64.8}
        & \grayc{44.8}
        \\
        & P2T-Tiny
        & 225
        & 44
        & 43.3
        & 65.7
        & 47.3
        & 39.6
        & 62.5
        & 42.3
        \\
        & \grayl{\textbf{P2T-Tiny+\short}} \vline
        & \grayc{232(+3.1\%)} 
        & \grayc{41(-6.8\%)} \vline
        & \grayc{\textbf{44.9}(+1.6)}
        & \grayc{67.0}
        & \grayc{49.1} \vline
        & \grayc{\textbf{40.7}(+1.1)}
        & \grayc{64.2}
        & \grayc{43.8}
        \\
        & Swin-T
        & 267
        & 41
        & 43.7
        & 66.6
        & 47.6
        & 39.8
        & 63.3
        & 42.7
        \\
        & \grayl{\textbf{Swin-T+\short}} \vline
        & \grayc{284(+6.4\%)} 
        & \grayc{38(-7.3\%)} \vline
        & \grayc{\textbf{44.3}(+0.6)}
        & \grayc{66.7}
        & \grayc{48.5} \vline
        & \grayc{\textbf{40.5}(+0.7)}
        & \grayc{63.6}
        & \grayc{43.4}
        \\
        & ConvNeXt-T$^{*}$
        & 262
        & 45
        & 43.8
        & 65.8
        & 48.2
        & 40.2
        & 62.9
        & 43.2
        \\
        & \grayl{\textbf{ConvNeXt-T+\short}} \vline
        & \grayc{279 (+6.5\%)} 
        & \grayc{42(-6.7\%)} \vline
        & \grayc{\textbf{45.6}(+1.8)}
        & \grayc{67.8}
        & \grayc{50.0} \vline
        & \grayc{\textbf{41.7}(+1.5)}
        & \grayc{64.9}
        & \grayc{44.8}
        \\
        \midrule
        Cascade
        & ResNet-50
        & 394
        & 99
        & 41.2
        & 59.4
        & 45.0
        & 35.9
        & 56.6
        & 38.4
        \\
        Mask R-CNN
        & \grayl{\textbf{ResNet-50+\short}} \vline
        & \grayc{405(+2.8\%)} 
        & \grayc{81(-18.2\%)} \vline
        & \grayc{\textbf{43.4}(+2.2)}
        & \grayc{62.3}
        & \grayc{47.2} \vline
        & \grayc{\textbf{37.7}(+1.8)}
        & \grayc{59.4}
        & \grayc{40.6}
        \\
        \bottomrule
        \toprule
        Method
        & Backbone
        & FLOPs (G)
        & FPS
        & $\text{AP}^{\text{b}}$
        & $\text{AP}^{\text{b}}_{50}$
        & \multicolumn{1}{c}{$\text{AP}^{\text{b}}_{75}$}
        & $\text{AP}^{\text{b}}_{S}$
        & $\text{AP}^{\text{b}}_{M}$
        & $\text{AP}^{\text{b}}_{L}$
        \\
        \midrule
        \multirow{2}{*}{RetinaNet}
        & ResNet-50
        & 239
        & 98
        & 36.5
        & 55.4
        & \multicolumn{1}{c}{39.1}
        & 20.4
        & 40.3
        & 48.1
        \\
        & \grayl{\textbf{ResNet-50+\short}} \vline
        & \grayc{253(+5.9\%)} 
        & \grayc{80(-18.4\%)} \vline
        & \grayc{\textbf{39.0}(+2.5)}
        & \grayc{58.9}
        & \grayc{41.7}
        & \grayc{23.6}
        & \grayc{43.0}
        & \grayc{51.4}
        \\
        \midrule
        Deformable 
        & ResNet-50
        & 173
        & 38
        & 44.5
        & 63.2
        & \multicolumn{1}{c}{48.9}
        & 28.0
        & 47.8
        & 58.8
        \\
        DETR
        & \grayl{\textbf{ResNet-50+\short}} \vline
        & \grayc{185(+6.9\%)} 
        & \grayc{34(-10.5\%)} \vline
        & \grayc{\textbf{45.6}(+1.1)}
        & \grayc{64.7}
        & \grayc{49.8}
        & \grayc{27.5}
        & \grayc{49.0}
        & \grayc{60.4}
        \\
        \bottomrule
    \end{tabular}
    }
    \caption{\textbf{COCO object detection and instance segmentation results}.
    $^{\dagger}$ indicates the backbone is re-trained by us. $^{*}$ indicates this reference model is trained by us because it is not provided by the reference.
    Mask R-CNN~\cite{maskrcnn2017}, Cascade Mask R-CNN~\cite{cascademaskrcnn2019}, and RetinaNet~\cite{retinanet2017} are trained with 1$\times$ schedule, while Deformable DETR~\cite{deformabledetr2020} is trained with 50-epoch schedule.
    Moreover, CoaT-based models are trained with multi-scale training inputs (MS)~\cite{swin2021,sparsercnn2021} following the CoaT~\cite{coat2021} training pipeline, while the other models are trained with single-scale training inputs (SS)~\cite{swin2021}. FLOPs are computed with an input resolution of 800$\times$1280. Throughputs are tested on an NVIDIA A100 GPU, and the other settings follow \cite{swin2021}.}
    \label{tab:detection}
\end{table*}

\noindent
\textbf{Settings} To evaluate the capability of vision backbones combined with our \meth~on object detection and instance segmentation, we conduct experiments on the COCO dataset~\cite{coco2014} with MMDetection~\cite{mmdet} toolbox, where the feature extractors are vision backbones integrated with \meth~pretrained on ImageNet-1k.
We first evaluate the performance of CoaT~\cite{coat2021}, ResNet~\cite{resnet2016}, PVTv2~\cite{pvtv22022}, P2T~\cite{p2t2022}, \red{swin~\cite{swin2021}, and ConvNeXt~\cite{liu2022convnet}} backbones on Mask R-CNN~\cite{maskrcnn2017}. Moreover, we test the performance of ResNet-50+\meth~with other mainstream detectors including Cascade Mask R-CNN~\cite{cascademaskrcnn2019}, RetinaNet~\cite{retinanet2017}, and Deformable-DETR~\cite{deformabledetr2020}. For ResNet ones, we use the default setting for ResNet from MMDetection with some minor changes. \red{For all other backbones, we apply their shared training configurations.} 
In general, all the detection networks except for Deformable DETR are trained for 12 epochs (1$\times$ schedule), while Deformable DETR is trained for 50 epochs. For CoaT-based models, we apply the multi-scale training strategy (MS)~\cite{swin2021,sparsercnn2021} following their original settings, while for the other models, we adopt the single-scale one (SS). 
The detailed training settings are in the Appendix.

\noindent
\textbf{Results} \cref{tab:detection} summarizes the performance of \meth~on several backbones. For CoaT, we replace its parallel blocks with MSCSA. CoaT Tiny+\meth~and CoaT Small+\meth~achieve better or similar accuracy with smaller FLOPs and significantly reduced runtime. This is because CoaT maintains parallel blocks in different resolutions and requires several rounds of communication between these parallel blocks. \red{Hence, replacing parallel blocks with the lightweight \meth~leads to improved computation efficiency.} 
For the other vision backbones, \meth~is applied as an add-on module, boosting their performance with less than \red{$7\%$} additional FLOPs. \red{Other than the CUDA-optimized ResNet50 backbones
\footnote{CUDA has specific runtime optimizations on convolutions, which are not available for the matrix multiplications in \meth. In Table \ref{tab:detection}, ResNet-50 with Mask R-CNN has a FLOPs of 260G, which is larger than or similar to the other backbones with Mask R-CNN. However, it has an FPS of 103, which is significantly larger.}
, the runtime increases in proportion to FLOPs.} 




\subsection{Semantic Segmentation}
\label{sec:seg}

\begin{table}[t]\small
    \centering
    \begin{tabular}{l|ccc}
        \toprule
        Backbone
        & FLOPs (G)
        & FPS
        & mIoU
        \\
        \midrule
        TopFormer-T$^{\dagger}$
        & 0.55
        & 960
        & 33.6 
        \\
        \grayl{\textbf{TopFormer-T+\short}} \vline
        & \grayc{0.55}
        & \grayc{948(-1.3\%)}
        & \grayc{\textbf{34.1}(+0.5)}
        \\
        TopFormer-B$^{\dagger}$
        & 1.8
        & 578
        & 38.3 
        \\
        \grayl{\textbf{TopFormer-B+\short}} \vline
        & \grayc{1.8}
        & \grayc{585(+1.2\%)}
        & \grayc{\textbf{38.9}(+0.6)}
        \\
        PVTv2-B0
        & 25
        & 305
        & 37.2
        \\
        \grayl{\textbf{PVTv2-B0+\short}} \vline
        & \grayc{26(+4.0\%)}
        & \grayc{285(-6.6\%)}
        & \grayc{\textbf{40.1}+(2.9)}
        \\
        PVTv2-B1
        & 34
        & 225
        & 42.5
        \\
        \grayl{\textbf{PVTv2-B1+\short}} \vline
        & \grayc{36(+5.9\%)}
        & \grayc{205(-8.8\%)}
        & \grayc{\textbf{44.1}(+1.6)}
        \\
        PVTv2-B2
        & 46
        & 146
        & 45.2
        \\
        \grayl{\textbf{PVTv2-B2+\short}} \vline
        & \grayc{49(+6.5\%)}
        & \grayc{132(-9.6\%)}
        & \grayc{\textbf{46.3}(+1.1)}
        \\
        \midrule
    \end{tabular}
    \caption{\textbf{ADE20k semantic segmentation results.}
    $^{\dagger}$ indicates the backbone of this reference model is re-trained by us.
    For TopFormer-based models, we use the segmentation head proposed by TopFormer~\cite{topformer2022}. The network is trained for 160k iterations with a batch size of 16. PVTv2-based models are trained with Semantic FPN~\cite{semanticfpn2019}, where the networks are trained for 40k iterations with a batch size of 32 following the shared pipeline. FLOPs are computed with 512$\times$512 resolution.}
    \label{tab:segmentation}
\end{table}

\noindent
\textbf{Settings} We also test vision backbones integrated with \meth~on semantic segmentation. We conduct experiments on TopFormer~\cite{topformer2022} and PVTv2~\cite{pvtv22022}-based models on the ADE20k dataset~\cite{ade20k2017} with MMSegmentation~\cite{mmseg} toolbox, where the backbones are pretrained on ImageNet-1k. For both TopFormer and PVTv2-based models, we use the corresponding training pipelines. And for TopFormer-based models, the segmentation head is the one proposed by TopFormer, whereas for PVTv2-based models, the models are trained with Semantic FPN~\cite{semanticfpn2019}. 
Following the configuration of each model, we train TopFormer ones with a mini-batch of 16 images for 160k iterations and PVTv2 ones with a mini-batch of 32 images for 40k iterations. 
The detailed training settings are in the Appendix.

\noindent
\textbf{Results} The semantic segmentation results are shown in~\cref{tab:segmentation}. For TopFormer backbones, \meth~replaces their Scale-Aware Semantics Extractor. \meth~improves mIoU and keeps the FLOPs and FPS to be nearly the same. For PVTv2 backbones, \meth~boosts the performance of PVTv2-B0/B1/B2 by 2.9\%/1.6\%/1.1\% with less than $10\%$ additional FLOPs and runtime. These results demonstrate the effectiveness of \meth~as an add-on module on semantic segmentation.

\subsection{Ablation Studies}
\label{sec:ablation}

This section describes a series of ablation studies on TopFormer-T~\cite{topformer2022}+\meth~on ImageNet-1k~\cite{imnet2009} to validate the effectiveness of each component in \meth. 
\begin{table}[h]\small
    \centering
    \begin{tabular}{l|ccc}
        \toprule
        Setting 
        & FLOPs (M)
        & FPS
        & Top-1 Acc.
        \\
        \midrule
        All opt. removed
        & 129
        & 9437
        & 65.1
        \\
        \midrule
        + w. PCP
        & 129(+0)
        & 9254(-1.9\%)
        & 65.7(+0.6)
        \\
        ++ w. Intra-FFN
        & 130(+1)
        & 9014(-4.5\%)
        & 66.0(+0.9)
        \\
        +++ w. MSP
        & 132(+3)
        & 8757(-7.2\%)
        & \textbf{67.1}(+2.0) 
        \\
        \bottomrule
    \end{tabular}
    \caption{\textbf{Component Analysis on TopFormer-T+\meth.} We start by removing all the component options from TopFormer-T+\meth. And the effectiveness of each major component is evaluated by cumulatively adding each component back.}
    \label{tab:ablation_major}
\end{table}

\begin{table}[h]
    \small
    \centering
    \begin{tabular}{l|ccc}
        \toprule
        Variants 
        & FLOPs (M)
        & FPS
        & Top-1 Acc.
        \\
        \midrule
        TopFormer-T+\short
        & 132
        & 8757
        & \textbf{67.1} 
        \\
        \midrule
        --/+ AvgPool
        & 132(-0)
        & 8744(-0.1\%)
        & 66.7(-0.4)
        \\
        --/+ Cas. Conv
        & 131(-1)
        & 8802(+0.5\%)
        & 66.9(-0.2)
        \\
        --/+ Sin. Conv
        & 131(-1)
        & 8931(+2.0\%)
        & 66.4(-0.7)
        \\
        \bottomrule
    \end{tabular}
    \caption{\textbf{Analysis of different down-sampling strategies.}
    \short~is \meth~in short. 
    The different design choices are illustrated in~\cref{fig:downsample}. The effectiveness of each design choice is evaluated on TopFormer-T+\meth~by replacing the default setting with the specific design.}
    \label{tab:ablation_downsample}
\end{table}
\label{sec:major_ablation}
\noindent
\textbf{Effectiveness of MSP, Intra-FFN and PCP}~\cite{hrvit2022}\textbf{.} The analysis of the major components is shown in~\cref{tab:ablation_major}, where the performance is evaluated by first removing all the components and then cumulatively adding each component back. The significant performance improvement comes from the Multi-Scale key, value tensor Projections (MSP) part, which brings about a 1.1\% accuracy gain with negligible additional FLOPs. This result confirms that the cross-scale feature interaction by MSP does help the model generalize better. On the other hand, by replacing half of the FFNs with our proposed Intra-FFN, we put more computational budget to increase the expansion ratio of linear projections in FFNs in the case of TopFormer. And this architectural modification gives us about a 0.3\% accuracy gain. Moreover, by adding the parallel convolution path (PCP)~\cite{hrvit2022} into the attention module, the top-1 accuracy increases by 0.6\%. This result validates the effectiveness of local feature aggregation in PCP.

\begin{figure}[!htb]
    \centering
    \begin{subfigure}{0.3\textwidth}
        \centering
        \includegraphics[width=0.98\textwidth]{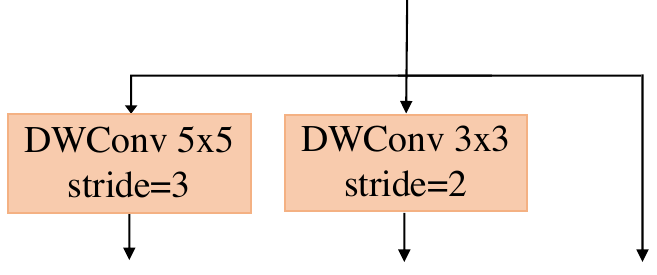}
        \caption{Default Setting.}
        \label{fig:default}
    \end{subfigure}
    \begin{subfigure}{0.3\textwidth}
        \centering
        \includegraphics[width=0.98\textwidth]{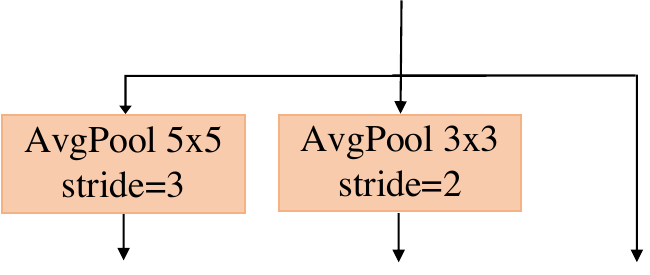}
        \caption{Average Pooling.}
        \label{fig:avgpool}
    \end{subfigure}
    
    \begin{subfigure}{0.3\textwidth}
        \centering
        \includegraphics[width=\textwidth]{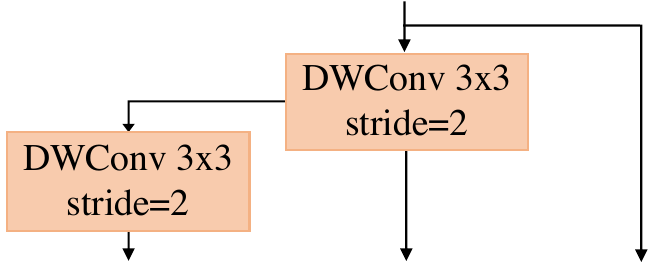}
        \caption{Cascade Convolution.}
        \label{fig:cascade}
    \end{subfigure}
    \begin{subfigure}{0.3\textwidth}
        \centering        
        \includegraphics[width=0.57\textwidth]{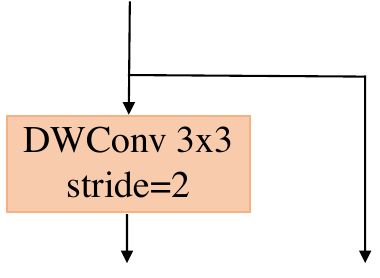}
        \caption{Single Convolution.}
        \label{fig:single}
    \end{subfigure}
    \caption{\textbf{Different down-sampling strategies.}}
    \label{fig:downsample}
\end{figure}

\label{sec:down-sampling}
\noindent
\textbf{Effectiveness of different down-sampling strategies in MSP.} Since the major contribution of our \meth~comes from the MSP part. Another series of ablation studies are conducted to evaluate the effectiveness of several down-sampling strategies. As illustrated in~\cref{fig:downsample}, instead of the default setting (\cref{fig:default}) in MSP, we replace the depthwise convolution with average pooling (\cref{fig:avgpool}), replace the parallel convolution structure with cascade convolutions (\cref{fig:cascade}), and drop the feature map in lowest resolution (\cref{fig:single}). Benefiting from more trainable parameters, the parallel convolution structure achieves the best performance as shown in~\cref{tab:ablation_downsample}, thus becoming our default setting in \attn. The performance gap between the single convolution design and the other three design choices demonstrates the advantage of generating feature maps in three different scales with small additional computational cost.


\section{Conclusion}

This work presented an add-on module, Multi-Stage Cross-Scale Attention (\meth), to enhance the performance of vision backbones by introducing interactions between feature maps in different stages and objects of different scales. \meth~works well for various vision backbones and attains a clear performance boost on common benchmark tasks with moderate additional FLOPs and runtime. 
\red{Future work involves applying \meth~on additional network architectures suitable for biomedical and neuroimaging tasks such as diffusion MRI pre-processing (e.g., denoising, distortion corrections) and estimation of biophysical and signal models, image registration, image quality transfer, stroke lesion segmentation, multi-modal brain age estimation, classification of disease and cognitive impairment status, prediction of progression to various forms of Alzheimer's and related dementias (ADRDs) such as the vascular contributions to cognitive impairments and dementia (VCID).}

\section{Acknowledgement}
We would like to acknowledge the support from NIH grants R01NS123378, P50HD105353, NIH R01NS105646, NIH R01NS11102, and R01NS117568.

\bibliographystyle{plain}
\bibliography{egbib}

\appendix

\section{Image Classification}
\label{sec:cls}

\noindent
\textbf{Settings} The performance on image classification is evaluated on \meth~with TopFormer~\cite{topformer2022}, CoaT~\cite{coat2021}, ResNet~\cite{resnet2016},  PVTv2~\cite{pvtv22022}, P2T~\cite{p2t2022}, Shunted~\cite{shunted2022}, Swin~\cite{swin2021}, and ConvNeXt~\cite{liu2022convnet} on the ImageNet-1k dataset~\cite{imnet2009}, 
where the implementation is based on PyTorch~\cite{pytorch2019} with MMCV~\cite{mmcv} and Timm~\cite{timm}. For a fair comparison, for CoaT, PVTv2, P2T, Shunted, Swin, and ConvNeXt backbones, we follow the same training pipelines as in their works. For the ResNet backbones, we adopt the training pipeline from PVTv2. Since TopFormer does not release their code for the training pipeline, we choose one from MobileNetV3~\cite{mobilenetv32019}, which is our best effort in reproducing their results. The detailed training settings are in~\cref{sec:cls_detail}.

\begin{table}[h]\small
    \centering
    \begin{tabular}{l|ccc}
        \toprule
        Model 
        & FLOPs (G)
        & FPS
        & Top-1 Acc.
        \\
        \midrule
        TopFormer-T$^{\dagger}$
        & 0.13
        & 9437
        & 65.1 
        \\
        \grayl{\textbf{TopFormer-T+\short}} \vline
        & \grayc{0.13}
        & \grayc{8757}
        & \grayc{\textbf{67.1}(+2.0)}
        \\
        TopFormer-B$^{\dagger}$
        & 0.38
        & 6746
        & 75.0 
        \\
        \grayl{\textbf{TopFormer-B+\short}} \vline
        & \grayc{0.38}
        & \grayc{5950}
        & \grayc{\textbf{75.5}(+0.5)}
        \\
        CoaT Tiny
        & 4.4
        & 120
        & 78.3
        \\
        \grayl{\textbf{CoaT Tiny+\short}} \vline
        & \grayc{4.4}
        & \grayc{432}
        & \grayc{\textbf{81.1}(+2.8)}
        \\
        CoaT Small
        & 13
        & 301
        & 82.1
        \\
        \grayl{\textbf{CoaT Small+\short}} \vline
        & \grayc{13}
        & \grayc{855}
        & \grayc{\textbf{83.0}(+0.9)}
        \\
        ResNet-50$^{\dagger}$
        & 4.1
        & 2404
        & 78.3 
        \\
        \grayl{\textbf{ResNet-50+\short}} \vline
        & \grayc{4.6}
        & \grayc{2050}
        & \grayc{\textbf{79.9}(+1.6)}
        \\
        PVTv2-B0
        & 0.57
        & 3797
        & 70.5
        \\
        \grayl{\textbf{PVTv2-B0+\short}} \vline
        & \grayc{0.65}
        & \grayc{3479}
        & \grayc{\textbf{74.6}(+4.1)}
        \\
        PVTv2-B1
        & 2.1
        & 2492
        & 78.7 
        \\
        \grayl{\textbf{PVTv2-B1+\short}} \vline
        & \grayc{2.3}
        & \grayc{2262}
        & \grayc{\textbf{80.0}(+1.3)}
        \\
        P2T-Tiny
        & 1.8
        & 2087
        & 79.8
        \\ 
        \grayl{\textbf{P2T-Tiny+\short}} \vline
        & \grayc{2.0}
        & \grayc{1957}
        & \grayc{\textbf{80.9}(+1.1)}
        \\
        Shunted-T
        & 2.1
        & 2103
        & 79.8
        \\
        \grayl{\textbf{Shunted-T+\short}} \vline
        & \grayc{2.3}
        & \grayc{1934}
        & \grayc{\textbf{81.0}(+1.2)}
        \\
        Swin-T
        & 4.4
        & 1578
        & 81.3
        \\
        \grayl{\textbf{Swin-T+\short}} \vline
        & \grayc{5.0}
        & \grayc{1414}
        & \grayc{81.5(+0.2)}
        \\
        ConvNeXt-T
        & 4.5
        & 1955
        & 82.1
        \\
        \grayl{\textbf{ConvNeXt-T+\short}} \vline
        & \grayc{5.1}
        & \grayc{1732}
        & \grayc{82.0(-0.1)}
        \\
        \bottomrule
    \end{tabular}
    \caption{\textbf{ImageNet-1k classification results.}
    $^{\dagger}$ indicates this reference model is re-trained by us.
    For TopFormer, we substitute the Scale-Aware Semantics Extractor with \meth. For CoaT, the parallel blocks are replaced by \meth. For all other backbones, \meth~is added in parallel to the backbones. For TopFormer and CoaT, we match the FLOPs of the corresponding reference model. For all other models, we set the FLOPs cap of \meth~to be around 10\% of the corresponding backbone. All models are trained with 224$\times$224 resolution. 
    }
    \label{tab:classification}
\end{table}




\noindent
\textbf{Results} The evaluation results of vision backbones combined with our \meth~are summarized in~\cref{tab:classification}. For TopFormer and CoaT, we replace their Scale-Aware Semantics Extractor and parallel blocks with our \meth, respectively. TopFormer+\meth~and CoaT+\meth~significantly surpass their corresponding baseline models with similar FLOPs. For CoaT, it maintains parallel blocks in different resolutions and requires several rounds of communication between these parallel blocks, while CoaT+\meth~replaces parallel blocks with \meth, leading to a much smaller runtime. For TopFormer backbones, the runtime is $7$-$8\%$ slower, likely due to increased memory usage in CSA and Intra-FFN. We believe that better optimizations in the parallel computations of CSA and Intra-FFN can resolve this.


We apply \meth~as an add-on module to CNN-based ResNet and ConvNeXt and ViT-based PVTv2, P2T, Shunted, and Swin backbones with moderate additional FLOPs and runtime (around 15\%). \meth~successfully boosts the performance of both CNN and ViT backbones. For Swin-T and ConvNeXt-T, \meth~does not affect the accuracy too much. However, as demonstrated in Tab. \ref{tab:detection}, \meth~brings a clear performance improvement on downstream tasks with Swin-T and ConvNeXt-T.



\section{Model Configuration}

\begin{table*}[t]
    \centering
    \resizebox{\textwidth}{!}{
    \begin{tabular}{l|cccccc}
        \toprule
        Backbone 
        & Pooling Size 
        & Squeeze Ratio  
        & Depth
        & \#Heads 
        & Head Dimension 
        & MLP Ratio
        \\
        \midrule
        TopFormer-T~\cite{topformer2022} 
        & 1/32 (1/64) 
        & - 
        & 2 
        & 4 
        & 14
        & 3 
        \\
        TopFormer-B~\cite{topformer2022}
        & 1/32 (1/64) 
        & - 
        & 2 
        & 8 
        & 12 
        & 3 
        \\
        \midrule
        CoaT Tiny~\cite{coat2021}
        & 1/16 (1/32)
        & - 
        & 3 
        & 8 
        & 22 
        & 1.5 
        \\
        CoaT Small~\cite{coat2021} 
        & 1/16 (1/32)
        & - 
        & 3 
        & 8 
        & 52 
        & 2 
        \\
        \midrule
        ResNet-18~\cite{resnet2016} 
        & 1/32 
        & 5/8 
        & 1 
        & 8 
        & 24 
        & 2 
        \\
        ResNet-50~\cite{resnet2016} 
        & 1/32 
        & 1/4 
        & 1 
        & 8 
        & 32 
        & 2 
        \\
        \midrule
        PVTv2-B0~\cite{pvtv22022} 
        & 1/32 
        & 3/4 
        & 1 
        & 8 
        & 15
        & 2
        \\
        PVTv2-B1~\cite{pvtv22022} 
        & 1/32 
        & 5/8 
        & 1 
        & 8 
        & 24
        & 2
        \\
        PVTv2-B2~\cite{pvtv22022} 
        & 1/32 
        & - 
        & 1 
        & 8
        & 35
        & 2
        \\
        \midrule
        P2T-Tiny~\cite{p2t2022} 
        & 1/32 
        & 3/4 
        & 1 
        & 8 
        & 36
        & 2
        \\
        \midrule
        Shunted-T~\cite{shunted2022} 
        & 1/32 
        & 5/8 
        & 1 
        & 8 
        & 30
        & 2
        \\
        \midrule
        Swin-T~\cite{swin2021} 
        & 1/32 
        & 3/4 
        & 1 
        & 12
        & 28
        & 2
        \\
        \midrule
        ConvNeXt-T~\cite{liu2022convnet} 
        & 1/32 
        & 3/4 
        & 1 
        & 12 
        & 28
        & 2
        \\
        \bottomrule
    \end{tabular}
    }
    \caption{\textbf{Model configuration for different backbones.}
    For TopFormer, we replace the Scale-Aware Semantics Extractor with \meth. For CoaT, the Parallel Blocks are replaced by \meth. For all other backbones, \meth~is added in parallel to the backbones. The pooling size for TopFormer follows the size used in ImageNet pre-training, while the value in parentheses is the size used in dense prediction tasks. For CoaT, we also use different pooling sizes for different tasks due to computational constraints.
    The dimension of value tensors (head dimension) is doubled from those of the query and key tensors, following LeViT~\cite{levit2021}.}
    \label{tab:model_config}
\end{table*}

To validate that our Multi-Stage Cross-Scale Attention (\meth) can enhance the performance of a wide variety of vision backbones via multi-stage and cross-attention feature map interactions, we implement and evaluate \meth~on several convolution-based and attention-based vision backbones. Our baselines include TopFormer~\cite{topformer2022} and CoaT~\cite{coat2021}, which have similar multi-stage feature interactions, where TopFormer has a convolution-based backbone, and CoaT has an attention-based backbone. We replace their attention modules and parallel blocks with \meth, respectively. Moreover, we test \meth~on the well-known CNN backbone ResNet~\cite{resnet2016} and ConvNeXt~\cite{liu2022convnet} and ViT backbone PVTv2~\cite{pvtv22022}, P2T~\cite{p2t2022},  Shunted~\cite{shunted2022}, and Swin~\cite{swin2021} to explore the performance gain of \meth~within a preset complexity cap, \ie, with an increase of approximately 10\% of the FLOPs of the backbone. Detailed model configurations are shown in~\cref{tab:model_config}. Parameters in the various configurations are:

\begin{itemize}
    \item Pooling Size: target pooling size at the beginning of \meth
    \item Squeeze Ratio: channel reduction ratio at the beginning of \meth, if applicable. 
    \item Depth: number of \meth~blocks used in \meth. 
    \item \#Heads: number of heads in \attn.
    \item Head Dimension: feature dimension of the query and key tensors in each head.
    \item MLP Ratio: the expansion ratio in FFN.
\end{itemize}

\section{Detailed Training Settings}

\subsection{ImageNet Classification}
\label{sec:cls_detail}

\noindent
\textbf{Implementation.}
All the models for ImageNet classification are implemented based on PyTorch~\cite{pytorch2019} with MMCV~\cite{mmcv} and Timm~\cite{timm}.

\noindent
\textbf{PVTv2}~\cite{pvtv22022} \textbf{and ResNet}~\cite{resnet2016}\textbf{.}
We adopt the shared training pipeline from PVTv2\footnote{\url{https://github.com/whai362/PVT}} and DeiT~\cite{deit2021} for PVTv2 and ResNet-based models, where the data augmentations include random cropping~\cite{cnn2017}, random horizontal flipping~\cite{cnn2017}, color jittering~\cite{cnn2017}, label smoothing~\cite{inceptionv32016}, random erasing~\cite{randomerasing2020}, repeated augmentation~\cite{repeataug2020}, AutoAugmentation~\cite{autoaug2018}, Mixup~\cite{mixup2017}, and CutMix~\cite{cutmix2019}. Moreover, the stochastic depth drop~\cite{stochasticdepth2016deep} is also applied to \meth~in PVTv2 and ResNet-based models. The models are trained for 300 epochs using AdamW~\cite{adamw2017} as the optimizer, with a weight decay of 0.05, a mini-batch of 1024 images, an initial learning rate of 0.001, 5 warm-up epochs, and a scaled cosine decay learning ratio scheduler.

\noindent
\textbf{CoaT}~\cite{coat2021}\textbf{.}
We adopt the shared training pipeline from CoaT\footnote{\url{https://github.com/mlpc-ucsd/CoaT}} and DeiT~\cite{deit2021} for CoaT-based models, where the data augmentations are identical to the ones used in PVTv2. Moreover, DropPath~\cite{droppath2016} is applied to \meth~in CoaT-based models following the design of parallel blocks in CoaT. The models are trained for 300 epochs using AdamW~\cite{adamw2017} as the optimizer, with a weight decay of 0.05, and a scaled cosine decay learning ratio scheduler. Furthermore, CoaT Tiny+\meth~is trained with a mini-batch of 2048 images, an initial learning rate of 0.002, 5 warm-up epochs, and exponential moving average (EMA), while CoaT Small+\meth~is trained with a mini-batch of 1024 images, an initial learning rate of 0.001, and 20 warm-up epochs.

\noindent
\textbf{TopFormer}~\cite{topformer2022}\textbf{.}
Since TopFormer has not released their code for the training pipeline at the time of this submission, we choose one from MobileNetV3~\cite{mobilenetv32019} and Timm\footnote{\url{https://huggingface.co/docs/timm/training_hparam_examples}}~\cite{timm}, which is our best effort in reproducing their image classification results. The data augmentations are of the same types as the ones used in PVTv2 and CoaT, but with some minor differences in numerical values. The stochastic depth drop~\cite{stochasticdepth2016deep} is applied to \meth~following the design of  the Scale-Aware Semantics Extractor in TopFormer. The model is trained for 600 epochs with RMSProp~\cite{rmsprop} as the optimizer, a weight decay of 1e-5, a mini-batch of 1024 images, 3 warm-up epochs, an initial learning rate of 0.064, a step learning rate scheduler with a decay epoch of 2.4, and EMA.

\noindent
\textbf{P2T}~\cite{p2t2022}\textbf{.}
We adopt the shared training pipeline from P2T\footnote{\url{https://github.com/yuhuan-wu/P2T}} and DeiT~\cite{deit2021} for P2T-based models, where the data augmentations are identical to the ones used in PVTv2. And the stochastic depth drop~\cite{stochasticdepth2016deep} is also applied to \meth~in P2T-based models. The models are trained for 300 epochs using AdamW~\cite{adamw2017} as the optimizer, with a weight decay of 0.05, a mini-batch of 1024 images, an initial learning rate of 0.001, 20 warm-up epochs, and a scaled cosine decay learning ratio scheduler.

\noindent
\textbf{Shunted}~\cite{shunted2022}\textbf{.}
We adopt the shared training pipeline from Shunted\footnote{\url{https://github.com/OliverRensu/Shunted-Transformer}} and DeiT~\cite{deit2021} for Shunted-based models, where the data augmentations are identical to the ones used in PVTv2. And the stochastic depth drop~\cite{stochasticdepth2016deep} is also applied to \meth~in Shunted-based models. The models are trained for 300 epochs using AdamW~\cite{adamw2017} as the optimizer, with a weight decay of 0.05, a mini-batch of 1024 images, an initial learning rate of 0.001, 5 warm-up epochs, and a scaled cosine decay learning ratio scheduler.

\noindent
\textbf{Swin}~\cite{swin2021}\textbf{.}
We adopt the shared training pipeline from Swin\footnote{\url{https://github.com/microsoft/Swin-Transformer}} and DeiT~\cite{deit2021} for Swin-based models, where the data augmentations are identical to the ones used in PVTv2. And the stochastic depth drop~\cite{stochasticdepth2016deep} is also applied to \meth~in Swin-based models. The models are trained for 300 epochs using AdamW~\cite{adamw2017} as the optimizer, with a weight decay of 0.05, a mini-batch of 1024 images, an initial learning rate of 0.001, 20 warm-up epochs, and a scaled cosine decay learning ratio scheduler.

\noindent
\textbf{ConvNeXt}~\cite{liu2022convnet}\textbf{.}
We adopt the shared training pipeline from Swin\footnote{\url{https://github.com/facebookresearch/ConvNeXt}} and DeiT~\cite{deit2021} for ConvNeXt-based models, where the data augmentations are identical to the ones used in PVTv2. And the stochastic depth drop~\cite{stochasticdepth2016deep} is also applied to \meth~in ConvNeXt-based models. The models are trained for 300 epochs using AdamW~\cite{adamw2017} as the optimizer, with a weight decay of 0.05, a mini-batch of 1024 images, an initial learning rate of 0.001, 20 warm-up epochs, and a scaled cosine decay learning ratio scheduler.

\subsection{COCO Detection and Instance Segmentation}

\noindent
\textbf{Implementation.}
All the models for COCO detection and instance segmentation are implemented based on MMDetection~\cite{mmdet}.

\noindent
\textbf{Mask R-CNN}~\cite{maskrcnn2017}\textbf{.}
For CoaT~\cite{coat2021}, PVTv2~\cite{pvtv22022}, and P2T~\cite{p2t2022}-based backbones, the detectors are trained with their shared training pipeline. For ResNet~\cite{resnet2016}-based backbones, the networks are trained with the default Mask R-CNN pipeline from the MMDetection~\cite{mmdet} toolbox. Specifically, we adopt the 1$\times$ schedule (12 epochs) to train all the models, which decreases the learning rate by 10$\times$ after epochs 8 and 11. While all the models use AdamW~\cite{adamw2017} as the optimizer with a mini-batch of 16 images and 500 linear warm-up iterations, CoaT-based models start with an initial learning rate of 0.0001 and a weight decay of 0.05, PVTv2 and ResNet-based models are trained with an initial learning rate of 0.0002 and a weight decay of 0.0001, and P2T-based models is trained with an initial learning rate of 0.0001 and a weight decay of 0.0001. Moreover, CoaT-based models utilize random horizontal flip~\cite{cnn2017} and multi-scale inputs (MS)~\cite{swin2021,sparsercnn2021} as data augmentation methods, where multi-scale inputs randomly resizes the shorter side of the image to be between 480 and 800 pixels while keeping the longer side of the image not exceeding 1333 pixels. PVTv2, P2T, and ResNet-based models are trained with random horizontal flip and single-scale inputs (SS)~\cite{swin2021}, where single-scale inputs resizes the shorter side of the image to be 800 pixels while keeping the longer side of the image not exceeding 1333 pixels.

\noindent
\textbf{RetinaNet}~\cite{retinanet2017}\textbf{and Cascade Mask R-CNN}~\cite{cascademaskrcnn2019}\textbf{.} We test ResNet+\meth~on RetinaNet and Cascade Mask R-CNN. On both detectors, we apply 1$\times$ schedule, random horizontal flip, and single-scale inputs. Moreover, the models are trained using SGD as the optimizer with a mini-batch of 16 images, an initial learning rate of 0.02, a momentum of 0.9, a weight decay of 0.0001, and 500 linear warm-up iterations.

\noindent
\textbf{Deformable DETR}~\cite{deformabledetr2020}\textbf{.} We also test ResNet+\meth~on Deformable DETR. Following its original setting, the models are trained for 50 epochs with random horizontal flip and multi-scale inputs while the learning rate decrease by 10$\times$ after epoch 40, with AdamW~\cite{adamw2017} as the optimizer, a mini-batch of 32 images, an initial learning rate of 0.0002, and a weight decay of 0.0001.

\subsection{ADK20k Segmentation}

\noindent
\textbf{Implementation.}
All the models for ADK20k Segmentation are implemented based on MMSegmentation~\cite{mmseg}.

\noindent
\textbf{TopFormer}~\cite{topformer2022}\textbf{.} Following the segmentation setting from TopFormer\footnote{\url{https://github.com/hustvl/TopFormer}}, the proposed segmentation head in TopFormer is applied. The models are trained for 160k iterations using AdamW~\cite{adamw2017} as the optimizer with a mini-batch of 16 images, an input size of 512 $\times$ 512, a weight decay of 0.01, a poly learning rate scheduler with a power of 1.0, and 1500 linear warm-up iterations. Moreover, the initial learning rate for TopFormer-T-based models is 0.0003, while the one for TopFormer-B-based modes is 0.00012. For data augmentation, random crop~\cite{cnn2017}, random horizontal flip~\cite{cnn2017}, and photometric distortions~\cite{cnn2017} are applied.

\noindent
\textbf{PVTv2}~\cite{pvtv22022} \textbf{.} PVTv2-based models utilize Semantic FPN~\cite{semanticfpn2019} for ADE20k segmentation. The models are trained for 40k iterations using AdamW~\cite{adamw2017} as the optimizer with a mini-batch of 32 images, an input size of 512 $\times$ 512 with the same data augmentation method as TopFormer-based models have, an initial learning rate of 0.0002, a weight decay of 0.0001, and a poly learning rate scheduler with a power of 0.9.

\section{Additional Implementaion Details}

\subsection{Multi-Stage Cross-Scale Attention (\meth) Blocks}

Our \meth~block follows the pre-norm~\cite{prenorm2019} setting that the normalization layer is applied to the input of every sub-layer. Recent work has demonstrated that removing the classification (CLS) token~\cite{bert2018,vit2020,deit2021} from attention blocks does not affect the performance~\cite{levit2021,visformer2021,pvt2021,hrvit2022}. Hence, during the design of the \meth~block, the CLS token is not applied and the overall attention block uses the BCHW tensor format for efficient implementations. For instance, all the Layer Normalization~\cite{layernorm2016} layers in \meth~blocks are replaced by the Batch Normalization~\cite{batchnorm2015} layers. All the linear projections are implemented by the 1$\times$1 convolutions.

\subsection{Cross-Scale Attention (\attn)}

The \attn~block applies the basic conv-attention setting of LeViT~\cite{levit2021}, reducing the number of channels in the query, key, and value tensors in order to save computation. The query and key tensors are further reduced to half the channel dimension of the value tensor. An additional Hardswish activation~\cite{mobilenetv32019} is also added to the attention output to encourage the nonlinear transformation in the attention module.

\noindent
\textbf{Parallel Convolution Path (PCP).}
The attention mechanism is good at modeling long-range relationships. However, it cannot distinguish between long-range and short-range relationships, where the structural information~\cite{structinfo2020} and local relationships~\cite{localrelation1999} play a significant role in vision tasks. Inspired by HRViT~\cite{hrvit2022}, we add a parallel convolution path between the value tensor in original resolution $\V_{0}$ and the output of self-attention $\text{Attention}(\Q, \K, \V)$ to preserve local information and act as the positional encoding. To be precise, the PCP module consists of a Hardswish~\cite{mobilenetv32019} activation and a 3$\times$3 depthwise convolution with padding size 1 and stride 1:
\begin{equation}
    \text{ConvPath}(\V_{0}) = \text{DWConv}(\text{Hardswish}(\V_{0})).
    \label{eq:conv_path}
\end{equation}
The Hardswish and depthwise convolution in PCP enable nonlinear inductive bias injection and local feature aggregation reinforcement. The modified self-attention now becomes
\begin{align}
    \begin{split} \text{ModifiedAttn}(\Q, \K, \V) = &\text{Attention}(\Q, \K, \V) + \text{ConvPath}(\V_{0}).
    \end{split}
\end{align}

\end{document}